\documentclass{ifacconf}

\usepackage{graphicx}

\usepackage{natbib}        % required for bibliography
\usepackage{times,fancyhdr,paralist}
\usepackage{amsmath,amssymb,amsfonts,mathrsfs}
\usepackage{amscd,bm}
\usepackage{amsxtra,dsfont,xspace,xfrac}
\usepackage{mathtools}

\newcommand{\fat}[1]{\mathds{#1}}

\newcommand{\RR}{\fat{R}}

\newcommand{\dd}[1]{\mathtt{d}#1}

\newcommand{\inv}{^{{\scriptscriptstyle -1}}}
\newcommand{\com}{^{{\ast}}}

\newcommand{\norm}[1]{\big\Vert #1\big\Vert}

\newcommand{\unitball}{\mathbb{B}}

\newcommand{\idm}[1]{{\mathbf{I}_{#1}}}

\newcommand{\expect}[2]{\mathbb{E}_{#1}\!\left[#2\right]}
\newcommand{\kldiv}[2]{D_{\!\scriptscriptstyle{K\!L}\!}\!\left(#1\Big\Vert #2\right)}

\newcommand{\work}{\Omega}
\newcommand{\ee}[1]{\mathtt{e}^{#1}}
\renewcommand{\ae}[1]{\text{a. e.}\left(#1\right)}

\newcommand{\goal}{x\com}
\newcommand{\upp}[1]{^{\scriptscriptstyle{(#1)}}}
\newcommand{\wgt}[2]{{\omega_{#1}\upp{#2}}}
\newcommand{\ts}[1]{{t_{#1}}}
\newcommand{\dts}[1]{\vartriangle\!\!t_{#1}}
\newcommand{\xhat}[2]{{\hat{x}_{#1}\upp{#2}}}

\newcommand{\intent}[2]{{\theta_{#1}\upp{#2}}}
\newcommand{\pt}[2]{{x_{#1}\upp{#2}}}
\newcommand{\rad}[2]{{r_{#1}\upp{#2}}}
\newcommand{\gt}[2]{{t_{#1}\upp{#2}}}
\newcommand{\ptusym}{f}
\newcommand{\assumed}[2]{\ptusym_{#1}\left(#2\right)}
\newcommand{\ptu}[2]{{\ptusym_{#1}\upp{#2}}}
\newcommand{\gain}[2]{{\lambda_{#1}\upp{#2}}}
\newcommand{\qrep}[2]{{q_{#1}\upp{#2}}}

\newcommand{\wdim}{n}

\newcommand{\rmin}{r_{\scriptscriptstyle{m\!i\!n}}}
\newcommand{\rmax}{r_{\scriptscriptstyle{m\!a\!x}}}

\newcommand{\tmin}{T_{\scriptscriptstyle{m\!i\!n}}}
\newcommand{\tmax}{T_{\scriptscriptstyle{m\!a\!x}}}

\newcommand{\effsym}{\scriptscriptstyle{e\!f\!f}}
\newcommand{\neff}{{N_{\effsym}}}
\newcommand{\wgteff}{{\omega_{\effsym}}}
\newcommand{\htheta}{{\hat{\theta}}}

\newcommand{\avgsym}{{\scriptscriptstyle{\mathtt{com}}}}
\newcommand{\redsym}{{\scriptscriptstyle{\mathtt{red}}}}
\newcommand{\hqred}{\hat{q}^\redsym}
\newcommand{\hqavg}{\hat{q}^\avgsym}
\newcommand{\hthavg}{\hat{\theta}^\avgsym}
\newcommand{\hthred}{\hat{\theta}^\redsym}

\newcommand{\errcov}{{\boldsymbol{\Delta}}}
\newcommand{\discov}{{\boldsymbol{\Sigma}}}
\newcommand{\pcov}[1]{{\mathbf{P}}_{#1}}
\newcommand{\kgain}[1]{{\mathbf{K}}_{#1}}

\newcommand{\hlow}{H_k^{low}}

\newcommand{\marg}[2]{{[#1]_{#2}}}

\newcommand{\major}[1]{\mathcal{M}_{#1}}
\newcommand{\allparticles}{\mathcal{P}}

\newcommand{\havg}[1]{H_{#1}^\avgsym}
\newcommand{\hred}[1]{H_{#1}^\redsym}

\newcommand{\havglow}[1]{\underline{H}_{{#1}}^\avgsym}
\newcommand{\hredlow}[1]{\underline{H}_{{#1}}^\redsym}

\newcommand{\revise}[1]{#1}
\begin{document}
\begin{frontmatter}

\title{Rao-Blackwellized particle filter for agent's intention inference\thanksref{number}}

\author[First]{Yixuan Wang} 
\author[Second]{Dan P. Guralnik} 
\author[First]{Warren E. Dixon}

\address[First]{Mechanical \& Aerospace Engineering Department, University of Florida.}
\address[Second]{Mathematics Department, Ohio University}

\thanks[number]{This research was supported in part by Air Force Office of Scientific Research award number FA9550-22-1-0429, FA9550-19-1-0169. Any opinions, findings and conclusions or recommendations expressed in this material are those of the author(s) and do not necessarily reflect the views of the sponsoring agency.}

\begin{abstract} 
Inferring the eventual goal of a mobile agent from noisy observations of its trajectory is a fundamental estimation problem.
We initiate the study of such intent inference using a variant of a Rao–Blackwellized Particle Filter (RBPF), subject to the assumption that the agent's intent manifests through closed-loop behavior with a state-of-the-art provable practical stability property.
%
%The closed-loop control gain $\lambda$ is selected using the Lyapunov envelope condition so that the tracking error decays exponentially under the assumed closed-loop dynamics.
%
Leveraging the assumed closed‑form agent dynamics, the RBPF analytically marginalizes the linear‑Gaussian substructure and updates particle weights only, improving sample efficiency over a standard particle filter.
Two difference estimators are introduced: a Gaussian mixture model using the RBPF weights and a reduced version confining the mixture to the effective sample.
We quantify how well the adversary can recover the agent’s intent using information‑theoretic leakage metrics and provide computable lower bounds on the Kullback–Leibler (KL) divergence between the true intent distribution and RBPF estimates via Gaussian‑mixture KL bounds.
We also provide a bound on the difference in performance between the two estimators, highlighting the fact that the reduced estimator performs almost as well as the complete one.
Experiments illustrate fast and accurate intent recovery for compliant agents, motivating future work on designing intent-obfuscating controllers.
\end{abstract}

\begin{keyword}
Estimation and filtering, Kalman filtering
\end{keyword}

\end{frontmatter}
%===============================================================================
\section{Introduction}
%
%\subsection{Motivation}\label{sec:intro subsec:motivation}
%
{\bf Motivation.} Anticipating an agent’s goals from partial, noisy observations is central to safe autonomy and human–robot interaction. 
Intent inference supports early decision making by allowing robots to reason over multiple plausible futures before conflicts arise, which is critical in navigation, assistance, and surveillance settings. 
\revise{The problem is intrinsically uncertain because the latent goal and arrival time induce distinct families of trajectories that often overlap in early observations.} 
%
%Robust intent inference therefore requires probabilistic models that capture discrete–continuous structure and deliver calibrated uncertainty in real time. 

Intent inference and obfuscation have been studied extensively in discretized environments, where the environment is modeled as a finite graph or Markov decision process \revise{(MDPs)} and high-level tasks are expressed in temporal logics \cite{kulkarni2018resource, Bernardini.ea2020, ijcai2017p610}.
In stochastic settings, entropy-maximizing policies and task-aware planning in MDPs have been used to minimize the predictability of trajectories or to limit the information leaked about high-level task specifications~\cite{Savas_2020,hibbard2019minimizinginformationleakageregarding}.
A complementary line of work leverages differential privacy to bound what an adversary can infer from observed paths, either by privatizing transition probabilities or by enforcing privacy at the level of symbolic trajectories and Markov chains~\cite{gohari2020privacypreservingpolicysynthesismarkov,8814723,chen2022differentialprivacysymbolicsystems}. 
These approaches typically operate on finite-state abstractions of the underlying dynamics and, apart from recent formulations of entropy-based obfuscation in partially observable models~\cite{Molloy_2023}, offer limited guidance on how to perform intent inference directly in continuous state spaces with nonlinear dynamics.

Particle filters propagate weighted hypotheses of latent variables under nonlinear, non-Gaussian dynamics, but their performance degrades as state dimension grows. 
RBPF~\cite{Doucet.RBPF.UAI.2000} addresses this difficulty by exploiting model structure: it samples a subset of the variables while analytically marginalizing the remainder using finite-dimensional optimal filters, thereby reducing variance and improving sample efficiency.
Its effectiveness has been demonstrated in tracking and \revise{simultaneous localization and mapping}~\cite{Arulampalam.Maskell.ea2002} as well as in multiple-target data association, where conditional linear–Gaussian substructures are common.
%
%Beyond tracking, RBPF has supported higher-level behavioral inference; for example, hierarchical transportation routines used RBPF at several abstraction levels to infer destinations and modes from GPS traces. 
%

%Despite these advances, intent inference methods frequently either assume black-box predictors that forgo model transparency or rely on generic particle filters that waste samples on continuous kinematics rather than on the intent variables that drive decision making. 
%

{\bf Contributions.} \revise{This work develops an RBPF formulation for agent intent inference.}
%of a Lyapunov-stabilized agent.
%
The model treats intent $\theta$ as a tuple (goal location $x$, goal radius $r$, arrival time $t$) and uses assumed closed-loop dynamics with bounded unknown disturbance to induce a conditionally linear–Gaussian evolution of the observed state.
Only the intent parameters are sampled, while analytically propagating the kinematic state. Particle weights are updated from trajectory observations via a Kalman update, concentrating the \revise{Monte Carlo} effort on the latent intent.

Two estimators representing intent as a distribution over the space of intent parameters are developed, representing weighted averages of the evolving set of particles, with and without pruning away ineffective particles.
A bound on the difference in information leakage between the two estimators is established.
Finally, numerical experiments indicate fast, accurate intent recovery in nominal conditions, with performance degrading gracefully under measurement noise. 

\section{Problem Formulation}\label{problem}
%
%A setting involving two principal entities in $\RR^\wdim$ with $\wdim\in\{2,3\}$, an adversary and an agent, is considered.
\revise{We consider two principal entities, an adversary and an agent, whose positions evolve in $\RR^\wdim$, $\wdim\in\{2,3\}$, corresponding to planar ($n=2$) or spatial ($n=3$) motion.
%
%The agent has the goal of reaching the ball $\goal+r\com\unitball$ within a time $t\com$.
%
The agent’s intent is specified by a triple $\theta\com\triangleq(\goal,r\com,t\com)$, where $\goal\in\RR^n$ is the goal center, 
$r\com > 0$ is the goal radius, and $t\com>0$ is the desired arrival time; the corresponding goal set is the ball\footnote{The symbol $\unitball$ henceforth denotes the Euclidean unit ball.} $\goal+r\com\unitball$, which the agent aims to reach by time $t\com$.
%
%The adversary’s objective is to infer the agent’s intent parameter $\theta\com\triangleq(\goal,r\com,t\com)$ by observing the agent’s trajectory.
%
The adversary’s objective is to infer the intent parameter $\theta\com$ by observing the agent’s trajectory.}
The adversary is assumed to be a passive observer, focusing the analysis on the agent's control decisions aimed at obscuring $\theta\com$ from the adversary.
%
%
%\subsection{Agent Dynamics and Task}\label{problem:agent}
%
%For simplicity, the agent is assumed to be fully actuated, but subject to bounded disturbances,
%
%\begin{align}\label{eqn:dynamics}
%    \dot{x}\in u+\bar{d}\unitball,
%\end{align}
%where the state $x\in\RR^\wdim$ is assumed to be measurable by the agent, $u$ is the control input, and $\bar{d}>0$ is a known constant.
%
%The task is encoded by the parameter $\theta\com$, whose
The values of $\theta\com$ are confined to the domain
\begin{equation}\label{eqn:intent domain}
    \Theta\triangleq R\unitball\times[\rmin,\rmax]\times[\tmin,\tmax],
\end{equation}
where $R>0$, $0<\rmin\ll\rmax\ll R$, and $0<\tmin\ll\tmax$ are parameters of the problem, fixed in advance.
\revise{Define mappings $x(\theta):\Theta\rightarrow\RR^n$, $r(\theta):\Theta\rightarrow\RR$ and $t(\theta):\Theta\rightarrow\RR$ such that, for each $\theta\in\Theta$, the quantities $x(\theta), r(\theta)$ and $t(\theta)$ denote the goal center, radius, and arrival time associated with intent $\theta$.
The notation $\theta=(x(\theta),r(\theta),t(\theta))$ for $\theta\in\Theta$ will be used when there is no risk of ambiguity.}
%

%For each $\theta\in\Theta$ and $q\in\RR^\wdim$, a smooth reference path
%\begin{align}
    %\label{eqn:reference paths}
    %&x_{q,\theta}:[0,\infty)\to\RR^\wdim, 
    %\label{eqn:reference paths conditions}
    %x_{q,\theta}(0)=q,  x_{q,\theta}(t(\theta))=x(\theta),
%\end{align}
%is available to the agent, whose task, given $x(0)=q$, is to track this path by selecting control inputs$u(x,t)$. %such that the tracking error $e(t)\triangleq x(t)-x_{q,\theta}(t)$ satisfies
%\begin{equation}\label{eqn:envelope constraint}
    %\norm{e(t)}\leq\varrho(t), 
    %\varrho(t(\theta))=r(\theta)
%\end{equation} 
%
%for some (TBD) increasing function $\varrho:[0,\infty)\to[0,\infty)$.\dg{ Should it be $\varrho_\theta$??}
%
% BRIEF DISCUSSION THAT THESE CONDITION ARE ENFORCEABLE (cite papers...) 
%
%
% MAYBE AN ILLUSTRATION HERE?
%
%In an unobstructed (homogeneous and isotropic) environment, it is natural to select
%\begin{equation}\label{eqn:reference path:interval}
    %x_{q,\theta}(t):=q+\tfrac{t}{t(\theta)}(x(\theta)-q)
%\end{equation}

%The agent is assumed to be aware of the presence of an adversary attempting to infer the intent parameter $\theta\com$ from its behavior.
%
%Therefore, %in addition to~\eqref{eqn:envelope constraint}, 
%the agent is tasked with selecting a controller whose resulting trajectory degrades the adversary's estimate $\htheta$ of the task parameter $\theta\com$ to an acceptable level within a time $\tacc<t(\theta)$, keeping in mind that the resulting bound $\varrho(t)$ needs to be kept as small as possible.
%

%\subsection{Adversary Capabilities and Task}\label{problem:adversary}
%
While the adversary may observe the agent and have knowledge of the agent dynamics%~\eqref{eqn:dynamics}
, the control input values deployed by the agent are \revise{typically unavailable} in real time to any outside observer.
However, information regarding the dependency of the controllers on the agent's task may be available, enabling the inference of the agent's goal from observations of its closed-loop behavior.
It is also reasonable for the adversary to assume that the closed loop dynamics of the agent, while possibly noisy, is designed to stabilize the goal within the allotted time.
Letting $\htheta$ denote the estimate of $\theta\com$ by the adversary, these considerations may be cast into the following assumption.

\begin{assum}\label{assumption: reasonable}
Given $\theta\in\Theta$, if $\htheta=\theta$ then the adversary estimates the closed-loop dynamics of the agent as
\begin{equation}\label{eqn:assumed dynamics}
    \begin{aligned}
        \dot{\hat{x}} &\in \underbrace{-\lambda(\theta)(\hat{x} - x(\theta))}_{\triangleq \assumed{\theta}{\hat{x}}} + \bar{d}\unitball,
    \end{aligned}
\end{equation}
where $\lambda:\Theta\to(0,\infty)$ is determined according to Lemma~\ref{lemma:stability of estimated controller}.
\end{assum}

The parameter $\bar{d}$ in~\eqref{eqn:assumed dynamics} represents an upper bound on the sum-total of unknown disturbances to the closed-loop dynamics $\ptusym_\theta$.
It is reasonable to assume this bound is known to the adversary, representing some knowledge of the agent's system capabilities in the environment.

\begin{lem}\label{lemma:stability of estimated controller} Let $\theta\in\Theta$ be a user-selected parameter and let
\begin{align}\label{eqn:gain requirements}
    \lambda_\theta \triangleq \max\left\{
        \frac{2\bar{d}}{r(\theta)},
        \frac{1}{t(\theta)}\log\frac{2R}{r(\theta)}
    \right\}.
\end{align}
Let $x\in\RR^\wdim$ evolve under the dynamics
\begin{equation}\label{eqn:coarsely linearized dynamics}
    \dot{x}\in -\lambda(x-x(\theta))+\bar{d}\unitball
\end{equation}
for some $\lambda\geq\lambda_\theta$.
Then, for any $x_0\in R\unitball$, the trajectory emanating from $x(0)=x_0$ satisfies $x(t)\in x(\theta)+r(\theta)\unitball$ for all $t\geq t(\theta)$.
\end{lem}
%
%\begin{proof} 
\begin{pf}
See Appendix~\ref{proof of lemma 1}.\qed
\end{pf}

%\end{proof}
%
Using the behavior model of the agent provided by~\eqref{eqn:assumed dynamics} and Lemma~\ref{lemma:stability of estimated controller} (specifically, with $\lambda(\theta)=\lambda_\theta$), the adversary seeks to infer the true intention $\theta\com$ from real-time observations of the agent's state.

\section{Intention Inference}\label{intention inference}
Unlike the standard PF, our work employs a RBPF to accelerate the Monte-Carlo process~\cite{best.Graeme.ea2015}, tailored specifically for the intention inference scenario.
Specifically, by Assumption~\ref{assumption: reasonable}, applying a Kalman Filter to the agent state-estimates corresponding to each particle instead of using the full Monte Carlo update reduces the computation load.
However, in the developed method, the particles will represent intent, rather than state.
Hence, they are not propagated through the filtering process, but the particle weights are updated at each timestep based on the discrepancy between the observed trajectory and the simulated trajectories generated by each particle.\footnote{Note that, as a result, a large number of particles is needed, to obtain a representative sample of $\Theta$.}

\subsection{RBPF Model Development}\label{intention inference:filtering}
%
%As the estimated intent parameter $\htheta$ could spread over the entire state space, it is natural to use PF to infer the intent parameter $\theta\com$.
%
\revise{We employe a set of $N\gg 1$ weighted particles $(\intent{k}{i},\wgt{k}{i})_{k=0,i=1}^{\infty,N}$ %are used 
to recursively estimate the intent parameter $\theta\com$  from incoming data.}
The initialization prior for the particles is uniform over $\Theta$.
Denote $\pt{k}{i}\triangleq x(\intent{k}{i})$, $\rad{k}{i}\triangleq r(\intent{k}{i})$, and $\gt{k}{i}\triangleq t(\intent{k}{i})$ for all $k\geq 0$ and $1\leq i\leq N$.
The adversary's noisy observation of the position of the agent at time $\ts{k}$ is modeled as
\begin{equation}\label{eqn:measurement model}
    y_k-x(\ts{k})\sim\mathcal{N}(0,\errcov)\nonumber,
\end{equation}
\revise{where $\mathcal{N}(0,\errcov)$ is the zero mean Gaussian distribution with symmetric positive-definite covariance $\errcov$.}
Each particle represents a possible evolution, $(\xhat{k}{i})_{k=0}^\infty$, of the agent's trajectory, given the assumed dynamics $\ptu{k}{i}\triangleq \assumed{\intent{k}{i}}{\xhat{k}{i}}$, provided in~\eqref{eqn:assumed dynamics}, with the associated gain $\gain{k}{i}\triangleq\lambda(\intent{k}{i})$.
These considerations motivate the following Bayesian filtering procedure.

{\bf Initialization.}
    The $N$ particles are assigned uniform weights, \revise{$\wgt{0}{i}\triangleq1/N$}, with the points $\pt{0}{i}$ sampled uniformly from the workspace $\work$; the radii $\rad{0}{i}$ sampled uniformly from a pre-defined interval $[\rmin,\rmax]$; and the times $\gt{0}{i}$ sampled uniformly from the pre-defined interval $[\tmin,\tmax]$.
    The initial state estimate corresponding to each particle $i$ is set to the initial measurement, $\xhat{0}{i}:=y_0$.

{\bf Propagation Step:}
    Discrete Euler integration of the dynamics~\eqref{eqn:assumed dynamics} provides a non-deterministic transition model\footnote{Note that this is the only place where $f_\theta$ comes to bear on the RBPF. Therefore, in principle, the developed framework may be applied for other assumed agent dynamics.} of particle $i$ for timestep $k$ yielding the prior $\xhat{k}{i-}$
   \begin{equation}\label{eqn:PF nondeterministic transition model}
        \xhat{k}{i-}
        -\xhat{k-1}{i}
        -\dts{k}\ptu{k-1}{i}
        %\gain{k-1}{i}\left(
        %    \xhat{k-1}{i}
        %    -\pt{k-1}{i}
        %\right)
        \in\dts{k}(\bar{d}\unitball)\revise{.}
    \end{equation}
    To obtain a probabilistic transition model, the disturbance term in~\eqref{eqn:PF nondeterministic transition model} is modeled as Gaussian noise
    \begin{equation}\label{eqn:PF probabilistic transition model}
        \xhat{k}{i-}
        -\xhat{k-1}{i}
        -\dts{k}\ptu{k-1}{i}
        %\gain{k-1}{i}\left(
        %    \xhat{k-1}{i}
        %    -\pt{k-1}{i}
        %\right)
        \sim\mathcal{N}(0,\dts{k}^2\discov)
    \end{equation}
    where $\discov\triangleq (\sigma\bar{d})^2\idm{\wdim}$ and $\sigma>0$ is a user-selected para\-meter determining the tail of the disturbance outside the ball $\bar{d}\unitball$.
    The prior error covariance $\pcov{k}^-$ then equals
    \begin{equation}\label{eqn:PF prior error covariance}
        \pcov{k}^-=(\gain{k-1}{i})^2\pcov{k-1} +\discov,
    \end{equation}
    and the Kalman gain $\kgain{k}$ is updated to
    \begin{align}\label{eqn:Kalman gain}
        \kgain{k} &=\pcov{k}^- 
        \left(
            \pcov{k}^- 
            +\errcov
        \right)\inv,
    \end{align}
    yielding the state posterior $\xhat{k}{i}$ and the posterior error covariance $\pcov{k}$
    \begin{equation}\label{eqn:posterior update}
    \begin{aligned}
        \xhat{k}{i} &= \xhat{k}{i-} + \kgain{k} (y_k - \xhat{k}{i-})\\
        \pcov{k} &= (I- \kgain{k})\pcov{k}^-.
    \end{aligned}
    \end{equation}

{\bf Update Step.}
The new measurement $y_k$ of the agent's position $x(\ts{k})$ causes an update of the particle weights \revise{as}
\begin{equation}\label{eqn:weight update}
\begin{aligned}
    & \wgt{k}{i}
    \propto
    \wgt{k-1}{i}p(y_k|\xhat{k}{i}),\\%\quad
    & p(y_k|\xhat{k}{i}) 
    \sim
    \mathcal{N}(\xhat{k}{i},\errcov),
\end{aligned}
\end{equation}
followed by a normalization of the weights.

{\bf Resampling Step.}\label{resampling step}
    Following a number of iterations, many particles could carry negligible weights, wasting computational resources.
To counter this problem, the effective sample size $\neff$ is defined by
\begin{equation}\label{eqn:effective sample size}
    \neff\triangleq \big\lfloor\left(
        \textstyle\sum_{i=1}^N (\wgt{k}{i})^2
    \right)^{-1}\big\rfloor.
\end{equation}
When $\neff$ drops below a predefined threshold $N_0$, we retain the $N_0$ highest weight particles and discard those with negligible weights.
The retained particles $(\intent{k}{i_a},\wgt{k}{i_a})_{a=1}^{N_0}$ are then replicated in proportion to their relative weights.
That is, each particle $(\intent{k}{i_a},\wgt{k}{i_a})$ is replaced with $N_a\triangleq\left\lfloor\wgt{k}{i_a}N\big/\sum_{b=1}^{N_0}\wgt{k}{i_b}\right\rfloor$ particles of weight $\frac{1}{N}$ and position $\intent{k}{i_a}$.
The remaining $N-\sum_{b=1}^{N_0}N_b$ particles are re-initialized in random positions in $\work$ with weights equal to $\frac{1}{N}$, to contribute to better coverage of the search space.

\begin{rem}\label{rem:intent particles do not move} It is important to note that $\intent{k+1}{i}=\intent{k}{i}$ unless the $i$-th particle is discarded during resampling.
In other words, the intent parameter attached to any particle is held constant throughout the particle's lifetime.
\end{rem}

\subsection{Estimation and Inference}\label{intention inference:estimation}
\begin{figure}[t]
    \centering
    
    \framebox{\parbox{3.3in}{\centering
    \includegraphics[width=3.3in]{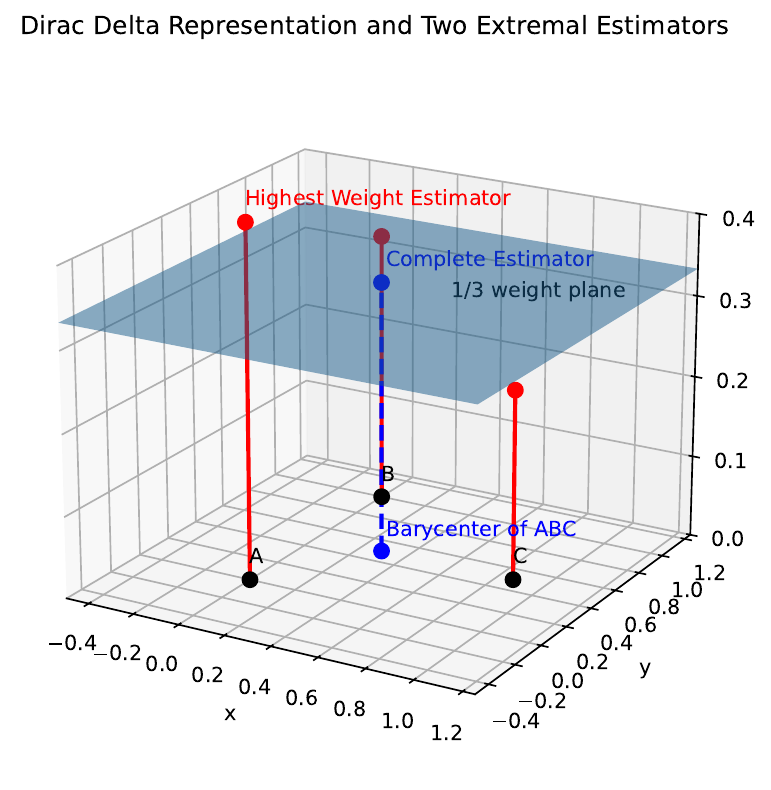}
}}
    \caption{\small Hypothetical situation with three particles estimating $x(\theta\com)$ with near-equal weights.
    The particle $A$ with weight $\tfrac{1}{3}+\varepsilon$, is a \revise{highest weight} estimate of $\theta\com$ according to the PF, but it represents a false positive with high probability ($\tfrac{2}{3}-\varepsilon$).
    At the same time, the weighted average of the particles may not be relevant at all, since its location is not contained in \emph{any} of the predicted goal regions.
    }
    \label{fig:Dirac representation}
\end{figure}
The weight update law of the RBPF may motivate the adoption of either the \revise{highest weight} particle, or the weighted average of the effective particles, as estimates of the agent's true goal, $\theta\com$.
This approach, however, is ineffectual for the purpose of inferring intent, as discussed in Figure~\ref{fig:Dirac representation}, calling for a more rigorous approach to estimation which enables quantifying the amount of information about $\theta\com$ gained from the estimator.

\subsubsection{Representation of Intent by a Distribution}\label{intention inference:estimation:intent densities}
At any point $t=\ts{k}$ in time, the $N$ weighted particles $(\intent{k}{i},\wgt{k}{i})_{i=1}^N$ may be used to give rise to estimators of $\theta\com$.
Any such estimator, ultimately, is a random variable, which may be regarded as a measurement of $\theta\com$.
Thus, each $\theta\in\Theta$ ought to be represented by a probability density $q_\theta$ over $\Theta$ with mean $\theta$.
Following our earlier convention, denote $\qrep{k}{i}\triangleq q_{\intent{k}{i}}$, for simplicity.
For computational simplicity, the $q_\theta$ are selected as continuous distributions\footnote{Returning to the initial considerations of Section~\ref{intention inference:estimation}, note also that selecting $q_\theta$ to be a Dirac distribution would not allow for evaluating the amount of information about $\theta\com$ present in $q_\theta$.
} over $\widetilde\Theta\triangleq\RR^\wdim\times(0,\infty)^2$, obtained as products of the form
\begin{equation}\label{eqn:probabilistic representation of intent}
    \begin{aligned}
        &q_\theta \sim \mathcal{N}(x(\theta),\sigma_x^2\idm{\wdim}) \otimes \mathcal{R}\otimes \mathcal{T},\\
        &\log(\mathcal{R}) = \mathcal{N}(r(\theta), \sigma_r^2),\\
        &\log(\mathcal{T}) = \mathcal{N}(t(\theta), \sigma_t^2),
    \end{aligned}
\end{equation}
where $\sigma_x,\sigma_r,\sigma_t$ are system parameters representing the capacity of the adversary for precise localization and timing and guaranteeing that an overwhelming probability is assigned to the event $\Theta\subset\widetilde\Theta$.

In computations, the marginal \revise{probability density functions} of $q_\theta$ will be denoted by $\marg{q_\theta}{x}$, $\marg{q_\theta}{r}$ and $\marg{q_\theta}{t}$, respectively.
\begin{figure}[t]
    \centering
    \framebox{\parbox{3.3in}{\centering
    \includegraphics[width=3.3in]{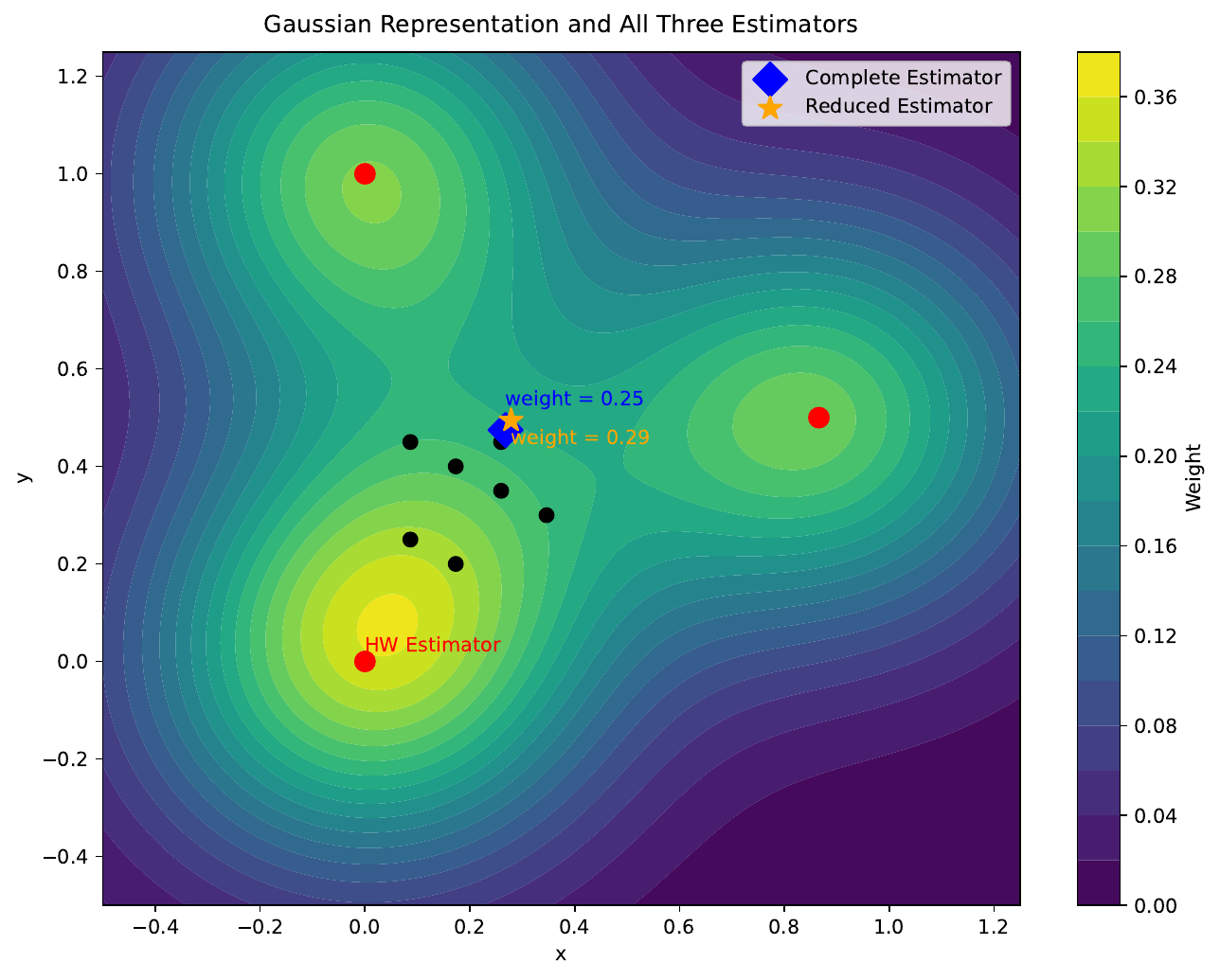}
}}
    \caption{\small Hypothetical situation with three effective particles with near-equal weights $\omega = 0.29$, represented by the reduced estimator (in distribution form), in Section~\ref{intention inference:estimation:effective estimator}.
    Seven additional particles with negligible weights are located near the barycenter of the three effective particles.
    Similar to Figure~\ref{fig:Dirac representation}, the \revise{highest weight} estimator remains a false positive with high probability.
    The position estimate from the weighted average of all particles \revise{$x(\hthavg_k)$} is distracted by the trivial weight particles, while the reduced estimator stays at the barycenter of effective particles with weight almost equal to the effective particles' weight.
    \revise{However, in} contrast to Figure~\ref{fig:Dirac representation}, either estimator provides information about the goal distribution $q_{\theta\com}$.
    }
      \label{fig:estimator choice}
   \end{figure}
\subsubsection{Density-Based Estimators of Intent}\label{intention inference:estimation:extremal estimators}
Replacing the PF with its associated collection of distributions $(\qrep{k}{i})_{i=1}^N$ allows for the construction of analogous, yet more meaningful estimators, $\htheta_k\triangleq\expect{}{\hat{q}_k}$, where the probability distributions $\hat{q}_k$ over $\widetilde\Theta$ are constructed as follows:

\noindent{\bf \revise{Highest Weight Estimator}.} The highest weight estimator corresponds to selecting $\hat{q}_k=\qrep{k}{i_0}$, where $i_0=\arg\max_i\wgt{k}{i}$.
    Note that it is now possible to quantify the discrepancy between $\theta\com$ and $\htheta_k$ using KL-divergence, as follows.
    Denote $\hat{x}_k \triangleq x(\htheta_k)$, $\hat{r}_k \triangleq r(\htheta_k)$, and $\hat{t}_k \triangleq t(\htheta_k)$ for all $k\geq 0$.
    \begin{figure}[t]
    \centering
    \framebox{\parbox{3.3in}{\centering
    \includegraphics[width=3.3in]{figures/line_last_frame.pdf}
}}
    \caption{\small
    \revise{RBPF intent–inference process.
    The upper-left panel shows the agent’s trajectory, and the upper-right panel plots the KL divergence for the complete and reduced estimators.
    The four lower panels display, from left to right, the effective sample size $\neff$, the position estimation error, the estimated goal radius, and the estimated arrival time.}
    }
      \label{fig:RBPF}
   \end{figure}
    Then,
    \begin{align}\label{eqn:KL divergence single particle}
        \textstyle\kldiv{q_{\theta\com}}{\hat{q}_k}
        &=
        \sum_{\nu\in\{x,r,t\}}\kldiv{\marg{q_{\theta\com}}{\nu}}{\marg{q_{\htheta_k}}{\nu}}\\
        &=
        \textstyle\frac{\norm{x\com-\hat{x}_k}^2}{2\sigma_x^2}
        +
        \frac{(r\com-\hat{r}_k)^2}{2\sigma_r^2}
        +
        \frac{(t\com - \hat{t}_k)^2}{2\sigma_t^2}.\nonumber
    \end{align}
    However, this estimator still suffers from the same drawback that was pointed out in Figure~\ref{fig:Dirac representation}.

\noindent{\bf Complete Estimator.} The challenges raised in Figure~\ref{fig:Dirac representation} are resolved through weighted averaging of the representative distributions of the particles (rather than the particles themselves), namely:
    \begin{equation}\label{eqn:weighted avg estimator}
        \hqavg_k\triangleq\sum_{i=1}^N \wgt{k}{i}\qrep{k}{i},
    \end{equation}
    noting that although $\htheta_k=\sum_{i=1}^N\wgt{k}{i}\intent{k}{i}$, the distribution $\hat{q}_k$ rarely satisfies $\hat{q}_k=q_{\htheta_k}$ (see Figure~\ref{fig:estimator choice}).
    This discrepancy emphasizes the fact that it is $\hat{q}_k$ that contains the most information about $\theta\com$ rather than just its first moment, motivating the study of bounds on $\kldiv{q_{\theta\com}}{\hat{q}_k}$.
    %
    %While this representation does capture all the information from the observation, notice from the figure \ref{fig:estimator choice} below,  the barycentric center of the particles does not represent the distribution well.
    %
    %It is reasonable to think the Barycentric center of the distribution should have some high probability to be the true intent distribution $q_{\htheta}$.
    %
    %However, as the origin distribution of $q_{\htheta}$ is a $\delta$ distribution, the spike will be sparse even in the current smoothed version, leading to the barycentric center's poor representation of the distribution.
%
\subsubsection{The Reduced Estimator}\label{intention inference:estimation:effective estimator}
Since ineffective particles are discarded in the resampling stage of the PF update, their contribution being minuscule, it is reasonable to maintain a reduced estimator of the form
\begin{equation}\label{eqn:reduced estimator}
    \begin{aligned}
        %q_{\htheta} &\sim \mathcal{X}\times \mathcal{R}\times \mathcal{T}\\
        %\mathcal{X} &=\sum_j \omega_j \mathcal{N}(x(\intent{k}{j}), \Sigma_x)\\
        %\mathcal{R} &= \sum_j \omega_j\textbf{Lognormal}(r(\intent{k}{j}), \sigma_r)\\
        %\mathcal{T} &= \sum_j \omega_j\textbf{Lognormal}(t(\intent{k}{j}), \sigma_t)
        \hqred_k\triangleq\sum_{i\in \major{k}}\frac{\wgt{k}{i}}{\wgteff}\qrep{k}{i},
    \end{aligned}    
\end{equation}
where the set $\major{k}\subseteq \allparticles$ is the set of %
%$j_1,\ldots,j_{\neff}$ are 
the $\neff$ highest weight particles \revise{and the set $\allparticles$ is }defined as the set of all particles, and $\wgteff=\sum_{i \in \major{k}} \wgt{k}{i}$.
%
%
%Instead of using all of the particles, using the top $\neff$ largest weight particles seems a perfect choice.
%
Figure~\ref{fig:estimator choice} gives an intuitive representation of the three estimators, from which \revise{the top $\neff$ representation contains the maximum information among all three.
%
%This gives the adversary rich capability to cross the information threshold by designing the CBF in Section~\ref{obfuscation}.
%
Then, $\hthavg_k $ and $ \hthred_k$ denote the intent estimates at time step $k$ defined as the expectations of the complete and reduced estimators, respectively.}
\section{Intent Estimators: Reduced vs. Complete}\label{comparison}
Let $\havg{k}$ and $\hred{k}$ denote the information leakage from the true intent distribution to the estimated one at time step $k$ using the average estimator and the reduced estimator, respectively,
%With the estimated intent distribution $q_\htheta$ at each time step and the true intent distribution $q_{\theta\com}$, we can now compute the KL-divergence between these two distributions at time step $k$, denoted as $H_k$, as the information leakage.
%
\begin{equation}\label{eqn:information leakage definition}
    \havg{k}\triangleq\kldiv{q_{\theta\com}}{\hqavg_k},\quad \hred{k}\triangleq\kldiv{q_{\theta\com}}{\hqred_k}.
\end{equation}
\revise{Since each $q_{\theta}$ is a product distribution over $\widetilde\Theta$}, see~\eqref{eqn:probabilistic representation of intent}, the KL divergences may be expanded as
\begin{equation}\label{eqn:H_k}
    \begin{aligned}
        \havg{k}
        &=
        \sum_{\nu\in\{x,r,t\}}
        \kldiv{\marg{q_{\theta\com}}{\nu}}{\textstyle\sum_{j \in \allparticles}\wgt{k}{j}\marg{\qrep{k}{j}}{\nu}}\\
        \hred{k}
        &=
        %\kldiv{\marg{q_{\theta\com}}{x}}{\mathcal{X}}+ \kldiv{\marg{q_{\theta\com}}{r}}{\mathcal{R}} + \kldiv{\marg{q_{\theta\com}}{t}}{\mathcal{T}},\\
        \sum_{\nu\in\{x,r,t\}}
        \kldiv{\marg{q_{\theta\com}}{\nu}}{\textstyle\sum_{i \in \major{k}}\tfrac{\wgt{k}{i}}{\wgteff}\marg{\qrep{k}{i}}{\nu}}.
    \end{aligned}
\end{equation}

\begin{thm}\label{thm:bound} The following lower bounds hold:
    \begin{equation}\label{Info lower bound}
    \begin{aligned}
        \havg{k} &\geq \havglow{k} \triangleq-\log\left(
        \sigma_x^{1-\wdim}(\frac{\ee{}}{2})^{\frac{\wdim+2}{2}}
        \right)\\
        &-\sum_{\nu\in\{x,r,t\}}\log\left(
        \sum_{j \in \allparticles} \wgt{k}{j}\ee{-\frac{\norm{\nu(\theta)- \nu(\qrep{k}{j})}^2}{4\sigma_{\nu}^2}}
        \right),\\
        \hred{k} &\geq \hredlow{k} \triangleq-\log\left(
        \frac{\sigma_x^{1-\wdim}}{\wgteff^3}(\frac{\ee{}}{2})^{\frac{\wdim+2}{2}}
        \right)\\
        &-\sum_{\nu\in\{x,r,t\}}\log\left(
        \sum_{i \in \major{k}} \wgt{k}{i}\ee{-\frac{\norm{\nu(\theta)- \nu(\qrep{k}{i})}^2}{4\sigma_{\nu}^2}}
        \right),
    \end{aligned}
\end{equation}
and there exist constants $C_x,C_r,C_t$ such that for all $k$:
    \begin{align}\label{eqn:thm1 bound}
    \Delta \hlow\triangleq
    |\havglow{k}-\hredlow{k}|
    \leq  \left|\sum_{\nu \in\{x,r,t\}}(\tfrac{1}{2}-\tfrac{1}{2\sqrt{2N}})\log C_\nu\right|.
\end{align}
\end{thm}
%\begin{proof}
\begin{pf}
    See Appendix \ref{app:proof 2}, where lower bounds on the $C_\nu$ are also derived, see Equation~\eqref{eqn:lower bound of dnu}.\qed
\end{pf} 
%
%\end{proof}
%

The lower bounds in~\eqref{Info lower bound} establish estimates of $\havg{k}$ and $\hred{k}$ in terms of data available to the adversary, in the absence of precise state information about the agent.
The discrepancy between the lower bounds serves as a conservative metric for the difference in the performance of the two respective estimators, as formalized in~\eqref{eqn:thm1 bound}.
%
%\dg{Talk about why $\Delta\hlow$ is important/interesting.}

\section{Numerical Experiments}
In this section, we evaluate the proposed RBPF for real-time intent inference in a planar goal-directed navigation task.
The primary objective is to quantify the speed and accuracy with which the adversary can recover the agent's hidden intent parameters $\theta^* = (x^*, r^*, t^*)$ using the derived density-based estimators.
\subsection{\revise{Simulation} Setup}
The experiment takes place in a 2D domain where the agent starts randomly at $x_0 \in R\unitball$ with a true goal $\theta^* \in \Theta$ such that $\norm{x^* - x_0}\geq 10$, $r^* \geq 1$ and $t^* \geq 20$.
The agent follows \revise{the stable closed-loop} dynamics \revise{in}~\eqref{eqn:assumed dynamics} with a disturbance bound $\revise{\bar{d}}=0.2$ and noise level $\sigma_d=0.38$.
The adversary employs an RBPF with $N=1200$ particles initialized uniformly over the workspace.
To rigorously measure the inference performance, we utilize the KL divergence in \eqref{eqn:H_k} as the primary metric of success.
We define Inference Time ($T_{in\!f}$) as the time $t_k$ at which the KL divergence drops below a threshold of $\epsilon = 50$ and remains bounded, indicating the adversary has successfully localized the intent distribution.

\subsection{Results and Analysis}
The simulation was repeated for $100$ Monte Carlo trials.
Figure \ref{fig:RBPF} illustrates the inference process of the RBPF.
Both estimators show an exponential decay in divergence as the agent moves toward the goal, reflecting the rapid reduction in uncertainty.
Table \ref{tab:performance} summarizes the terminal accuracy and inference speed, from which the Reduced Estimator consistently maintains lower final error states by pruning the ineffective particles.
This \revise{result }indicates that maintaining the complete estimator offers no information advantage over the reduced estimator, except possibly for rare cases (which are, at any rate, accounted for by the information leakage lower bound in Theorem \ref{thm:bound}).
\begin{table}[t]
    %\centering
    \begin{tabular}{|l|c|c|}
        \hline
        Metric & Reduced Estimator & Complete Estimator \\
        \hline
        \hline
        Inference Time & $3.28s\pm 2.28$ & $3.28s\pm 2.28$\\
        Final Position Est. Error & $0.225m\pm0.295$ & $0.231m\pm0.295$\\
        Final Radius Est. Error & $0.264m\pm0.636$ &  $0.274m\pm0.626$\\
        Final Time Est. Error & $0.729s\pm0.699$ & $0.732s\pm0.697$\\
        \hline
    \end{tabular}\\[.5em]
    \caption{\small Performance comparison of intent estimators with 100 times repeat experiments.}
    \label{tab:performance}
\end{table}

At the very beginning of a trajectory, the posterior distribution is diffuse, with particles covering both candidate goals and exhibiting high variance.
As the agent commits to a direction likelihood updates collapse the distribution around the correct goal, leading to rapid posterior concentration.

Across all runs, $\neff$ initially falls as weights become uneven, then rises after resampling, and oscillates during late-stage convergence.
The position estimation error monotonically decreases as the particle cloud contracts around the true state trajectory.
The predicted arrival-time histograms tighten as uncertainty shrinks, reflecting improved confidence in the inferred goal.
The RBPF identifies the correct goal well before the halfway point of the trajectory, and the final estimate matches the ground truth with negligible error.
\section{Conclusion}
Intent inference for goal‑directed motion as an RBPF over $(x,r,t)$ were implemented, exploiting analytic Kalman updates to marginalize the linear–Gaussian part of the dynamics and thereby improving sample efficiency relative to standard particle filters.
Density‑based estimators of intent were developed, and the information leakage was evaluated using computable lower bounds to the KL divergence between the true intent distribution and estimator mixtures.
%, adopting efficient bounds for Gaussian‑vs‑Gaussian mixture model comparisons. %from~\cite{Durrieu.Thiran.2012}. 
%
\revise{Simulations} demonstrate fast, accurate intent recovery despite noise.
The framework separates control from inference and yields diagnostics with KL‑bounds, and posterior marginals that are reproducible across tasks.
Future work includes richer motion models and non‑Gaussian observation channels, principled design of obfuscation controllers that minimize KL‑based leakage subject to task constraints while keeping track of reference path, and extensions to multi‑agent settings.

\begin{ack} The authors thank the reviewers for their careful reading of the manuscript and for their constructive suggestions, which substantially improved the clarity and rigor of our presentation.
\end{ack}

\bibliography{ifacconf}  

\appendix

\section{Proof of Lemma~\ref{lemma:stability of estimated controller}}\label{proof of lemma 1}
Let $e\triangleq x-\revise{x(\theta)}$ denote the error state and consider the Lyapunov function candidate $V(e)\triangleq\norm{e}^2$.
Following~\cite[Theorem 10.1.3]{Aubin2008}, let $x:[0,T]\to\RR^\wdim$, $T\in(0,\infty]$ be a complete solution of the differential inclusion~\eqref{eqn:coarsely linearized dynamics}.
Applying the chain rule
%~\cite[\dg{Proposition YYY}]{Aubin2008} 
to the composition $V\circ e$ and noting that $V$ is a smooth function yields
\begin{align}
    %\nonumber
    \dot{V}&\in 
    2e^\top(-\lambda e+\bar{d}\unitball)=
    -2\lambda V
    +2\bar{d}e^\top\unitball,
\end{align}
for almost all $t$.
Hence,
\begin{equation}    \label{eqn:lyapunov inequality}
    \dot{V}\leq -2\lambda V+2\bar{d}V^{\frac{1}{2}}
     \ae{[0,T]}.
\end{equation}
Multiplying both sides of~\eqref{eqn:lyapunov inequality} by $\tfrac{1}{2}V^{-1/2}\exp(\lambda t)$ and integrating yields, after some algebra:
\begin{align}
%    &&\tfrac{1}{2}V^{-1/2}\dot{V}\ee{\lambda t}
%    +\lambda V^{\frac{1}{2}}\ee{\lambda t}
%    &\leq
%    \bar{d}\ee{\lambda t}\nonumber\\
%    &\iff&\frac{d}{dt}\left(V^{1/2}\ee{\lambda t}\right) &\leq \bar{d}\ee{\lambda t}\nonumber\\
%    &\iff&\left(V^{1/2}\ee{\lambda t}\right)\Big|_{0}^t &\leq \frac{\bar{d}}{\lambda}\ee{\lambda t}\Big|_{0}^t\nonumber\\
    V(e(t))^\frac{1}{2} \leq V(e(0))^{1/2}\ee{-\lambda t} &+ \frac{\bar{d}}{\lambda}(1 - \ee{-\lambda t})
\end{align}
Using $\norm{e(0)}\leq R$ and $V^\frac{1}{2}=\norm{e}$ results in
\begin{align}\label{eqn:error bound}
    \norm{e}&\leq R\ee{-\lambda t} + \frac{\bar{d}}{\lambda} (1- \ee{-\lambda t}).
\end{align}
%As time $t$ grows, the error $e$ will converge to $\frac{\bar{d}}{\lambda}$.
%
%At the same time, by selecting an arbitrarily large control gain $\lambda$, the error norm $\norm{e}$ will be arbitrarily small.
%
\newcommand{\rev}[1]{{\textcolor{red}{#1}}}
To ensure $\norm{e}\leq r(\theta)$ for all $t > t(\theta)$, it is required that:
\begin{align}\label{ineqn:error-disturbance}
    R\ee{-\lambda t} + \frac{\bar{d}}{\lambda} \left(
        1-\ee{-\lambda t}
    \right) 
    \leq 
    r(\theta),
\end{align}
which is equivalent to
\begin{align}\label{eqn:gain condition 1}
    \lambda\left(
        1-\frac{R}{r(\theta)}\ee{-\lambda t}
        \right)
        &\geq
        \frac{\bar{d}}{r(\theta)}\left(
            1-\ee{-\lambda t}
        \right),
\end{align}
It follows from~\eqref{eqn:gain requirements} and $\lambda\geq\lambda_\theta$ that the second factor on the left-hand side of~\eqref{eqn:gain condition 1} is positive---in fact, greater than $\tfrac{1}{2}$---for all $t\geq t(\theta)$.
Therefore, for all such $t$, \eqref{eqn:gain condition 1} is equivalent to
\begin{align}\label{eqn:gain condition 2}
    \lambda &\geq
    \frac{\bar{d}}{r(\theta)} \frac{1 - \ee{-\lambda t}}{1 - \frac{R}{r(\theta)}\ee{-\lambda t}}.
\end{align}
Observing that $1-\ee{-\lambda t}\leq 1$ and that $1-\tfrac{R}{r(\theta)}\ee{-\lambda t}\geq\tfrac{1}{2}$
for all $t\geq t(\theta)$, one concludes it suffices for $\lambda$ to satisfy  $\lambda\geq\tfrac{2\bar{d}}{r(\theta)}$, which is also provided by~\ref{eqn:gain requirements}.\hfill\qed

\section{Proof of Theorem \ref{thm:bound}}\label{app:proof 2}
Applying lower bounds for KL-divergences between a single Guassian and a Gaussian mixture model (GMM) from~\cite{Durrieu.Thiran.2012} yields lower bounds for each term.
For example, a lower bound for the $x$ term in the reduced estimator of~\eqref{eqn:H_k} is
\begin{equation}\label{eqn:KLdiv x lower bound}
    \begin{aligned}
        &\kldiv{\marg{q_{\theta\com}}{x}}{\textstyle\sum_{i \in \major{k}}\tfrac{\wgt{k}{i}}{\wgteff}\marg{\qrep{k}{i}}{x}}\\
        &\geq
        -\log \left(
        \sum_i \wgt{k}{i}\ee{-\frac{||x(\theta)- x(\qrep{k}{i})||^2}{4\sigma_x^2}}
        \right)
        -\log\left(\frac{\sigma_x^{1-\wdim}}{\wgteff}\frac{\ee{}}{2}^{\frac{\wdim}{2}}
        \right).
    \end{aligned}
\end{equation}
\revise{where $n$ denotes the dimension of the position $x(\theta)$; in this paper, as stated in Section \ref{problem}, we restrict attention to $n\in{2,3}$.}
%
%In this paper, as stated in Section~\ref{problem}, we set $n=2,3$.
%
%\begin{equation}\label{eqn:kldiv derivation 1}
    %\begin{aligned}
        %&\kldiv{\marg{q_{\theta\com}}{x}}{\textstyle\sum_{i \in \major{k}}\tfrac{\wgt{k}{i}}{\wgteff}\marg{\qrep{k}{i}}{x}}\\
        %&\geq 
        %-\log\left(
            %\sum_i\textstyle
            %\frac{\wgt{k}{i}}{\wgteff} %\int_{\widetilde{\Theta}} 
            %\marg{q_{\theta\com}}{x}
            %\marg{\qrep{k}{i}}{x}
            %\dd{x}
        %\right) 
        %-\frac{1}{2}\log\left(
        %(2\pi\ee{})^\wdim(\sigma_x)^2
        %\right).
    %\end{aligned}
%\end{equation}
%The probability density functions of $\marg{q_{\theta\com}}{x}$ and $\marg{\qrep{k}{i}}{x}$ are:
%\begin{align*}
    %\marg{q_{\theta\com}}{x} &= \frac{1}{(2\pi\sigma_x^2)^{\wdim/2}}\ee{-\frac{\norm{x-x(\theta)}^2}{2\sigma_x^2}}\\
    %\marg{\qrep{k}{i}}{x} &= \frac{1}{(2\pi\sigma_x^2)^{\wdim/2}}\ee{-\frac{\norm{x-x(\qrep{k}{i})}^2}{2\sigma_x^2}}
%\end{align*}
%Then the integral is:
%
%\begin{align*}
    %&\int \marg{q_{\theta\com}}{x} \marg{\qrep{k}{i}}{x}\dd{x}=
    %\frac{\ee{-\frac{\norm{x(\theta)- x(\qrep{k}{i})}^2}{4\sigma_x^2}}}{(4\pi\sigma_x^2)^{\wdim/2}},
%\end{align*}
%
%
\if false
\begin{align*}
    &\int \marg{q_{\theta\com}}{x} \marg{\qrep{k}{i}}{x}\dd{x}\\
    &= \int \frac{1}{(2\pi\sigma_x^2)^{\wdim/2}}\ee{-\frac{\norm{x-x(\theta)}^2}{2\sigma_x^2}} \frac{1}{(2\pi\sigma_x^2)^{\wdim/2}}\ee{-\frac{\norm{x-x(\qrep{k}{i})}^2}{2\sigma_x^2}}\dd{x}\\
    %&= \frac{1}{2\pi\sigma_x^2}\int\ee{-\frac{(x-x(\theta))^2 +(x-x(\qrep{}{i}))^2}{2\sigma_x^2}} dx\\
    %&= \frac{1}{2\pi\sigma_x^2}\int\ee{-\frac{2(x-\frac{x(\theta)+ x(\qrep{}{i})}{2})^2 +\frac{(x(\theta)- x(\qrep{}{i}))^2}{2}}{2\sigma_x^2}} dx\\
    &=
    \frac{\ee{-\frac{\norm{x(\theta)- x(\qrep{k}{i})}^2}{4\sigma_x^2}}}{(2\pi\sigma_x^2)^\wdim}\int\ee{-\frac{\norm{x-\frac{x(\theta)+ x(\qrep{k}{i})}{2}}^2}{\sigma_x^2}}\dd{x}\\
    &=
    \frac{\ee{-\frac{\norm{x(\theta)- x(\qrep{k}{i})}^2}{4\sigma_x^2}}}{(2\pi\sigma_x^2)^\wdim}\cdot
    \left(
        2\pi(\tfrac{\sigma_x}{\sqrt{2}})^2
    \right)^{n/2}\\
    &=
    \frac{\ee{-\frac{\norm{x(\theta)- x(\qrep{k}{i})}^2}{4\sigma_x^2}}}{(4\pi\sigma_x^2)^{\wdim/2}},
\end{align*}
\fi
%
%
%because of the identity
%\begin{align*}
    %\int \ee{-\frac{\norm{x-\mu}^2}{2\sigma^2}}\dd{x} = (2\pi\sigma^2)^\frac{\wdim}{2}.
%\end{align*}
%Substituting these results into~\eqref{eqn:kldiv derivation 1}, one has:
%
%\begin{equation}
    %\begin{aligned}
        %&\kldiv{\marg{q_{\theta\com}}{x}}{\textstyle\sum_{i \in \major{k}}\tfrac{\wgt{k}{i}}{\wgteff}\marg{\qrep{k}{i}}{x}}\\
        %&\geq
        %-\log\left(
        %\sum_i \frac{\wgt{k}{i}\ee{-\frac{||x(\theta)- x(\qrep{k}{i})||^2}{4\sigma_x^2}}}{\wgteff(4\pi\sigma_x^2)^{\wdim/2}}
        %\right)
        %-\frac{1}{2}\log\left(
        %(2\pi\ee{})^\wdim(\sigma_x)^2
        %\right)\\
        %&=-\log \left(
        %\sum_i \wgt{k}{i}\ee{-\frac{||x(\theta)- x(\qrep{k}{i})||^2}{4\sigma_x^2}}
        %\right)
        %-\log\left(\frac{\sigma_x^{1-\wdim}}{\wgteff}\frac{\ee{}}{2}^{\frac{\wdim}{2}}
        %\right).
    %\end{aligned}
%\end{equation}
%

%
Collecting all three terms together amounts to the lower bound $\havglow{k}$ and $\hredlow{k}$.
%, where
%\begin{equation}\label{Info lower bound}
%    \begin{aligned}
%        \havglow{k} \triangleq&-\log\left(
%        \sigma_x^{1-\wdim}(\frac{\ee{}}{2})^{\frac{\wdim+2}{2}}
%        \right)\\
%        &-\sum_{\nu\in\{x,r,t\}}\log\left(
%        \sum_{j \in \allparticles} \wgt{k}{j}\ee{-\frac{\norm{\nu(\theta)- \nu(\qrep{k}{j})}^2}{4\sigma_{\nu}^2}}
%       \right)\\
%        \hredlow{k} \triangleq&-\log\left(
%        \frac{\sigma_x^{1-\wdim}}{\wgteff^3}(\frac{\ee{}}{2})^{\frac{\wdim+2}{2}}
%        \right)\\
%        &-\sum_{\nu\in\{x,r,t\}}\log\left(
%        \sum_{i \in \major{k}} \wgt{k}{i}\ee{-\frac{\norm{\nu(\theta)- \nu(\qrep{k}{i})}^2}{4\sigma_{\nu}^2}}
%        \right)
%    \end{aligned}
%\end{equation}
%
Upper bounds similar to \eqref{Info lower bound} exist, from which a convergence analysis of the information can be derived in future work.
%
%$\havg{k}\leq \havgup{k}$ and $\hred{k}\leq \hredup{k}$, where
%\begin{equation}\label{Info upper bound}
%    \begin{aligned}
%        \havgup{k} \triangleq&\log\left(
%        \sigma_x^{1-\wdim}(\frac{\ee{}}{2})^{\frac{\wdim+2}{2}}
%        \right)\\
%        &-\sum_{\nu\in\{x,r,t\}}\log\left(
%        \sum_{j \in \allparticles} \wgt{k}{j}\ee{-\frac{\norm{\nu(\theta)- \nu(\qrep{k}{j})}^2}{2\sigma_{\nu}^2}}
%        \right)\\
%        \hredup{k} \triangleq&\log\left(
%        \sigma_x^{1-\wdim}\wgteff^3(\frac{\ee{}}{2})^{\frac{\wdim+2}{2}}
%        \right)\\
%        &-\sum_{\nu\in\{x,r,t\}}\log\left(
%        \sum_{i \in \major{k}} \wgt{k}{i}\ee{-\frac{\norm{\nu(\theta)- \nu(\qrep{k}{i})}^2}{2\sigma_{\nu}^2}}
%        \right)
%    \end{aligned}
%\end{equation}
%
%Then the difference between the upper and lower bounds of information leakage for the average estimators is:
%
%\begin{align*}
    %\Delta H_k^{\avgsym} &\triangleq \mid\havgup{k} - \havglow{k}\mid\\
    %&= 2\log\left(
        %\sigma_x^{1-\wdim}(\frac{\ee{}}{2})^{\frac{\wdim+2}{2}}
        %\right)\\
        %&+\sum_{\nu\in\{x,r,t\}}\log\left(\frac{\sum_{j \in \allparticles} \wgt{k}{j}\ee{-\frac{\norm{\nu(\theta)- \nu(\qrep{k}{j})}^2}{4\sigma_{\nu}^2}}}{\sum_{j \in \allparticles} \wgt{k}{j}\ee{-\frac{\norm{\nu(\theta)- \nu(\qrep{k}{j})}^2}{2\sigma_{\nu}^2}}}
        %\right)
%\end{align*}
%We also measure the information leakage difference between the two estimators.
%
%For each time step $k$, the additional information $\Delta H_k\triangleq \havg{k} - \hred{k}$ provided by the ineffective particles introduces the change to the information lower bound $\Delta \hlow$:
%
We proceed to analyze $\Delta \hlow$
\begin{align}
    &\Delta \hlow=\mid\havglow{k} - \hredlow{k}\mid\nonumber\\
    &=\left|\sum_{\nu \in \{x,r,t\}}
        \log\left ( 
        \wgteff\frac{\sum_{j\in\allparticles}\wgt{k}{j}\ee{-\frac{\norm{\nu(\theta) - \nu(\qrep{k}{j})}^2}{4\sigma_{\nu}^2}}}{\sum_{i \in \major{k}}\wgt{k}{i}\ee{-\frac{\norm{\nu(\theta) - \nu(\qrep{k}{i})}^2}{4\sigma_{\nu}^2}}}
    \right )\right|,
\end{align}
\revise{Define the function $f(\nu)$ and $d_\nu$ as follows}:
\begin{align}
    f(\nu) &\triangleq \frac{\sum_{j\in\allparticles \setminus \major{k}}\wgt{k}{j}d_\nu}{\sum_{i \in \major{k}}\wgt{k}{i}d_\nu}, \quad
    d_\nu \triangleq \ee{-\frac{\norm{\nu(\theta) - \nu(\qrep{k}{j})}^2}{4\sigma_{\nu}^2}}
\end{align}
The constraint $\theta\in\Theta$ implies:
\begin{align}\label{eqn:dnu bound}
    d_x \geq \ee{-\frac{R^2}{\sigma_{\nu}^2}},\;
    d_r \geq \ee{-\frac{\norm{\rmax - \rmin}^2}{4\sigma_{\nu}^2}},\;
    d_t \geq \ee{-\frac{\norm{\tmax - \tmin}^2}{4\sigma_{\nu}^2}},
\end{align}
which motivates defining
\begin{align}\label{eqn:lower bound of dnu}
    \underline{d}_x \triangleq\ee{-\frac{R^2}{\sigma_{\nu}^2}},\;
    \underline{d}_r\triangleq\ee{-\frac{\norm{\rmax - \rmin}^2}{4\sigma_{\nu}^2}},\;
    \underline{d}_t\triangleq\ee{-\frac{\norm{\tmax - \tmin}^2}{4\sigma_{\nu}^2}}.
\end{align}
Therefore, an upper bound on $f(\nu)$ is derived 
\begin{align}
    f(\nu)= \frac{\sum_{j\in\allparticles \setminus \major{k}}\wgt{k}{j}d_\nu}{\sum_{i \in \major{k}}\wgt{k}{i}d_\nu}
    \leq \frac{\sum_{j\in\allparticles \setminus \major{k}}\wgt{k}{j}}{C_\nu \cdot \sum_{i \in \major{k}}\wgt{k}{i}}
    = \frac{1-\wgteff}{C_\nu\wgteff},
\end{align}
where $C_\nu \triangleq\min_{i\in\major{k}} d_\nu \in [\underline{d}_\nu, 1]$.
Similarly,
\begin{align}
    f(\nu)\geq \frac{1-\wgteff}{\wgteff} B_\nu,
\end{align}
where $B_\nu \triangleq\min_{i\in\allparticles\setminus\major{k}} d_\nu \in [\underline{d}_\nu, 1]$.
Then the information difference $\Delta \hlow$ is upper bounded by:
\begin{align}
    \Delta \hlow &=\left|\sum_{\nu \in\{x,r,t\}}\log\left(\wgteff(1+ f(\nu))\right)\right|\nonumber\\
    %\Delta H_k &\leq \left|\sum_{\nu \in\{x,r,t\}}\log(\wgteff + \frac{1-\wgteff}{C_\nu})\right|\\
    %&\leq
    %\left|\sum_{\nu \in\{x,r,t\}}\left(
        %\wgteff\log(1)
        %+(1-\wgteff)\log\frac{1}{C_\nu}
    %\right)\right|\\
    &\leq
    \left|\sum_{\nu \in\{x,r,t\}}(1-\wgteff)\log C_\nu\right|,
\end{align}
and the information difference $\Delta \hlow$ is lower bounded by:
\begin{align}
    \Delta \hlow &\geq \left| \sum_{\nu \in\{x,r,t\}}\log(\wgteff + (1-\wgteff)B_\nu)\right|.
\end{align}
To finish the proof, recall an earlier result:
\begin{prop}[\protect{\cite[Section 4.2]{wang2025effectivemodelpruning}}]\label{prop:point on boundary} If $\neff=1$ then $\wgteff\geq\tfrac{1}{2}$.
If $\neff=N$ then $\wgteff=1$.
Otherwise, 
\begin{align}\label{eqn:phi nu inf result}
    \wgteff\geq\frac{\neff}{N}+\frac{N-\neff}{N}\sqrt{\frac{N-\neff-1}{(\neff+1)(N-1)}}.
\end{align}
The bounds are sharp.
\end{prop}

\revise{Now, applying Proposition~\ref{prop:point on boundary} to $\wgteff$, the bounds for the effective weight become $\wgteff \in [(1 + \tfrac{1}{\sqrt{2N}})/2, 1]$.}
Hence, the information difference $\Delta \hlow$ can be bounded by:
\begin{align}
    0 \leq \Delta \hlow \leq  \left|\sum_{\nu \in\{x,r,t\}}(\frac{1}{2}-\frac{1}{2\sqrt{2N}})\log C_\nu\right|,
\end{align}
as claimed.\hfill\qed
\end{document}